\documentclass[letterpaper, 10 pt, conference]{ieeeconf}  

\IEEEoverridecommandlockouts                              

\usepackage{graphics} 
\usepackage{epsfig} 
\usepackage{mathptmx} 
\usepackage{times} 
\usepackage{amsmath} 
\usepackage{amssymb}  
\usepackage{bbm}
\usepackage{dsfont}
\usepackage[dvipsnames]{xcolor}
\usepackage[utf8]{inputenc}
\usepackage{subcaption}
\usepackage{pdfpages}
\usepackage{paralist}
\usepackage[font=small,labelfont=bf]{caption}
\usepackage{array}
\usepackage{booktabs}
\usepackage{lipsum}
\usepackage{caption}
\usepackage{siunitx}
\usepackage{cite}
\usepackage{mathtools}
\usepackage{xcolor}
\usepackage{blkarray, bigstrut}
\usepackage{svg}
\let\labelindent\relax
\usepackage{enumitem}
\usepackage[bookmarksopen, bookmarksdepth=2, breaklinks=true ]{hyperref}
\RequirePackage{color,graphicx}
\definecolor{linkcolour}{rgb}{0.2,0.285,0.918}
\hypersetup{colorlinks,linkcolor={blue},citecolor={linkcolour},urlcolor={blue}} 
\usepackage{graphicx}
\usepackage{multirow}
\usepackage[dvipsnames]{xcolor}

\AtBeginEnvironment{quote}{\raggedright}

\usepackage{authblk}

\title{Is the House Ready For Sleeptime? \\Generating and Evaluating Situational Queries for Embodied Question Answering \vspace{-0.3cm}}

\author[$1, 2$]{Vishnu Sashank Dorbala}
\author[$2$]{Prasoon Goyal}
\author[$2$]{Robinson Piramuthu}
\author[$2$]{Michael Johnston}
\author[$1$]{\\ Reza Ghanadan}
\author[$1$]{Dinesh Manocha\vspace{-0.2cm}}
\affil[$1$]{University of Maryland, College Park}
\affil[$2$]{Amazon AGI}
\affil[ ]{\small{Supplementary materials can be found at \url{https://gamma.umd.edu/seqa/}}}

\newcolumntype{M}[1]{>{\centering\arraybackslash}m{#1}}

\newcommand\Item[1][]{%
  \ifx\relax#1\relax  \item \else \item[#1] \fi
  \abovedisplayskip=0pt\abovedisplayshortskip=0pt~\vspace*{-\baselineskip}}

\begin{document}

\maketitle
\thispagestyle{empty}
\pagestyle{empty}

\begin{abstract}
We present and tackle the problem of Embodied Question Answering (EQA) with \textit{Situational Queries} (S-EQA) in a household environment. Unlike prior EQA work tackling simple queries that directly reference target objects and properties (``What is the color of the car?"), situational queries (such as ``Is the house ready for sleeptime?") are challenging as they require the agent to correctly identify multiple object-states (Doors: Closed, Lights: Off, etc.) and reach a \textit{consensus} on their states for an answer.
Towards this objective, we first introduce a novel Prompt-Generate-Evaluate (PGE) scheme that wraps around an LLM's output to generate \textit{unique} situational queries and corresponding consensus object information. PGE is used to generate $2K$ datapoints in the VirtualHome simulator, which is then annotated for ground truth answers via a large scale user-study conducted on M-Turk. With a high rate of answerability ($97.26\%$) on this study, we establish that LLMs are good at generating situational data.
However, in evaluating the data using an LLM, we observe a low correlation of $46.2\%$ with the ground truth human annotations; indicating that while LLMs are good at generating situational data, they struggle to answer them according to consensus. When asked for reasoning, we observe the LLM often goes against commonsense in justifying its answer. Finally, we utilize PGE to generate situational data in a real-world environment, exposing LLM hallucination in generating reliable object-states when a structured scene graph is unavailable.
To the best of our knowledge, this is the first work to introduce EQA in the context of \textit{situational queries} and also the first to present a \textit{generative} approach for query creation. We aim to foster research on improving the real-world usability of embodied agents through this work.

\vspace{-0.1cm}
\end{abstract}

\begin{figure}[t!]
    \centering
    \includegraphics[width=0.95\linewidth]{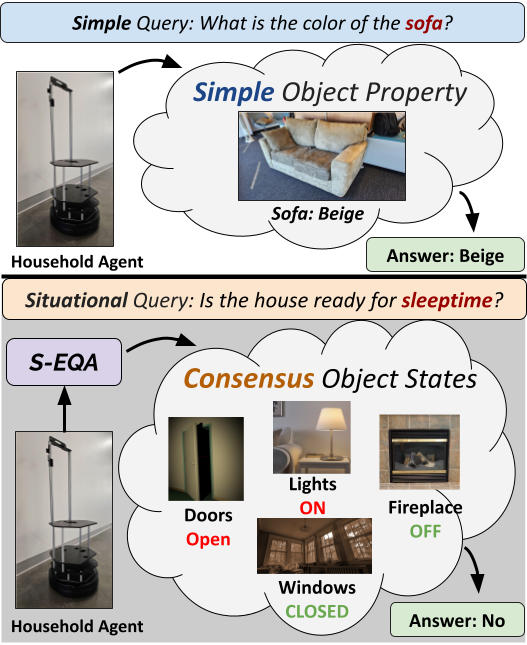}
    \caption{\textbf{Simple} vs \textbf{Situational EQA}: We introduce \textit{Situational Embodied Question Answering} (S-EQA), where queries require inspecting multiple objects and deriving \textit{consensus} knowledge of their states. Using an LLM, we generate S-EQA on VirtualHome \cite{virtualhome}, with \textit{consensus} object states and relationships. A large-scale MTurk study annotates and verifies data \textit{authenticity}. Evaluating LLMs on S-EQA reveals their strength in generating queries and consensus but misalignment in answering them. LLM explanations further suggests poor commonsense reasoning on our task. 
    }
    \label{fig:cover}
    \vspace{-0.9cm}
\end{figure}

\section{Introduction}
\label{sec: Intro}

Embodied Question Answering (EQA) introduced by Das. et. al., \cite{eqa} involves an agent exploring its surroundings to find an answer to a query. Prior work on this task \cite{eqa2, eqa3, mteqa, iqa} has primarily dealt with what we term ``\textit{simple queries}", that are directly answerable from states or properties of objects from the simulator. For instance, questions like ``What is the color of the sofa?" or ``Is there an object nearby to help me cut vegetables?" are directly related to an object present in the environment, requiring a mix of vision-language grounding with elementary commonsense to solve the task. While important, this type of question-answering capability does not add much utility to a real-world household robot, as the user asking questions is likely to already know answers to simple questions about their surroundings.

\begin{table*}[t!]
\centering
\setlength{\tabcolsep}{8pt} 
\renewcommand{\arraystretch}{1.5} 
\begin{tabular}{>{\centering\arraybackslash}m{2cm} >{\centering\arraybackslash}m{2.5cm} >{\centering\arraybackslash}m{1.5cm} >{\centering\arraybackslash}m{1.5cm} >{\centering\arraybackslash}m{1.5cm} >{\centering\arraybackslash}m{2.5cm} >{\centering\arraybackslash}m{2cm}}
\hline
\textbf{Dataset} & \textbf{Navigation + QA} & \textbf{Multi-Target} & \textbf{Interaction} & \textbf{Abstract Queries} & \textbf{Object State Consensus} & \textbf{Query Creation} \\ \hline
EQA \cite{eqa} & \checkmark & - & - & - & - & \textit{Rule-Based} \\ 
MT-EQA \cite{mteqa} & \checkmark & \checkmark & - & - & - & \textit{Rule-Based} \\ 
IQA \cite{iqa} & \checkmark & \checkmark & \checkmark & - & - & \textit{Rule-Based} \\ 
K-EQA \cite{knowledge-eqa} & \checkmark & \checkmark & - & \checkmark & - & \textit{Rule-Based} \\ 
S-EQA \textbf{(Ours)} & \checkmark & \checkmark & - & \checkmark & \checkmark & \textbf{\textit{Generative}} \\ \hline
\end{tabular}
\caption{\textbf{Comparison of S-EQA with existing EQA datasets}. We introduce \textbf{situational queries} that require \textit{consensus} object states for answers. While K-EQA asks abstract questions (e.g., “Is there an \underline{object} used to cut food nearby?” → \textit{Yes} if a \underline{\textit{knife}} is present), S-EQA extends this by requiring \textit{consensus} on multiple object states (e.g., “Is the kitchen ready for meal prep?” → \textit{Knives: Present, Oven: Preheated, etc.}). As real-world consensus data is scarce, we approximate it using an LLM and validate authenticity via a user study. We are the first to introduce a \textbf{generative} approach for EQA dataset creation, and discuss challenges associated with such schemes.}
\label{table:methods_comparison}
\vspace{-0.5cm}
\end{table*}

In contrast, answering queries such as ``\textit{Is the kitchen ready for meal preparation?}" requires a sense of situational awareness in gauging the status of multiple objects in the environment. While not uniformly defined, a \textit{consensus} does exist on standard conditions that can be used to answer such questions. This refers to a set of generally accepted object states and conditions (such as 
a \underline{Knife} being \underline{Present}) for successfully evaluating the query.  
A household robot deployed in the real world is likely to encounter such queries that require it to utilize consensus knowledge pertaining which objects and associated states are commonly associated with the situation. We term these as \textbf{\textit{situational queries}}, and in this work, tackle the twofold challenge of 1) conjuring \textit{authentic} situational queries and corresponding consensus object data and 2) highlighting the role of consensus object states in simplifying the process of responding to situational queries.

A query such as ``\textit{Is the house ready for sleeptime?}" is a \textit{situational query}, since there exist generally accepted object states, such as dimmed lights, closed windows, locked doors, etc., which most people associate with a house being ready for nighttime.
However, a query like ``\textit{Can I bring guests over?}", while situational, is too \textit{personal}, as the answer relies on various user preferences on household norms and event-specific constraints. In our work, we aim to generate and human-validate queries where a general \textit{consensus} exists on multiple object states.

Developing EQA datasets in the past has been limited to constructing simple queries using entities from a simulator \cite{mteqa, iqa}, with some work extending it to abstract queries built on scene-graphs \cite{knowledge-eqa}. Posing \textit{situational queries} using these existing techniques is however nontrivial, as they stem from human intuition and contextual awareness of the environment. Setting up real-world experiments to collect this data is also cumbersome; if not having an intrusive agent capturing conversations at home, an offline approach would be cognitively heavy, requiring humans to imagine situations for queries. In our work, we develop a scheme that utilizes a Large Language Model (LLM) to generate such queries, providing in-context cures to cover unique concepts stemming from various situations. Table \ref{table:methods_comparison} compares our dataset with existing work.

\noindent
\textbf{Main Results}: We tackle the issues presented above, utilizing LLMs as a prior for generating situational queries and associated consensus object states and relationships\footnote{\href{https://youtu.be/m3hg59zGYg8}{https://youtu.be/m3hg59zGYg8}}.
We human-validate the generated data, and present a benchmark for EQA on situational queries. The main contributions of our work are as follows:- 

\begin{itemize}
    \item \textbf{Problem}: We tackle the novel problem of \textbf{\textit{posing}} and evaluating \textbf{\textit{Situational Queries}} for Embodied Question Answering (\textbf{S-EQA}). Unlike simple queries that directly reference particular objects, posing situational queries itself is challenging, requiring a \textit{consensus} on multiple objects and their states, and cannot be formed easily using prior rule-based methods (Refer Table \ref{table:methods_comparison}). 
    \item \textbf{Method}: We propose a novel Prompt-Generate-Evaluate (\textbf{PGE}) scheme that integrates an LLM with a feedback scheme that allows for the generation of several \textbf{unique} situational queries, 
    along with corresponding \textit{consensus} object-state and object-relationship information that needs to be satisfied to successfully answer the question.
    \item \textbf{Human-Validation}: Using PGE in the VirtualHome simulator, we  produce a S-EQA dataset of $2000$ situational queries and consensus data. We annotate these via a large scale MTurk \cite{mturk} user study, which gives us a high authenticity percentage of $97.26\%$, indicating that the data generated corresponds to consensus.
    \item \textbf{LLM Eval}: We prompt the LLM to provide binary answers to S-EQA datapoints generated from VirtualHome. Despite being able to generate highly authentic situational data, we observe a subpar correlation of $46.2\%$ with the human annotations when it comes to answering. We infer that LLMs can \textbf{generate good situational queries} and consensus object-states, but \textbf{answer them poorly}. 
    
    \item \textbf{LLM Reasoning}: When asked for an explanation, we observe the \textbf{poor commonsense reasoning of LLMs} in justifying their answer. This further suggests that LLMs exhibit poor situational awareness.
    \item \textbf{Real World Challenges}: Finally, we discuss the transfer of PGE to a real world real world lab environment, and discuss various challenges with LLM hallucination that obstruct the generation of quality consensus data.
\end{itemize}

\begin{figure*}[t!]
    \centering
    \includegraphics[width=0.8\linewidth]{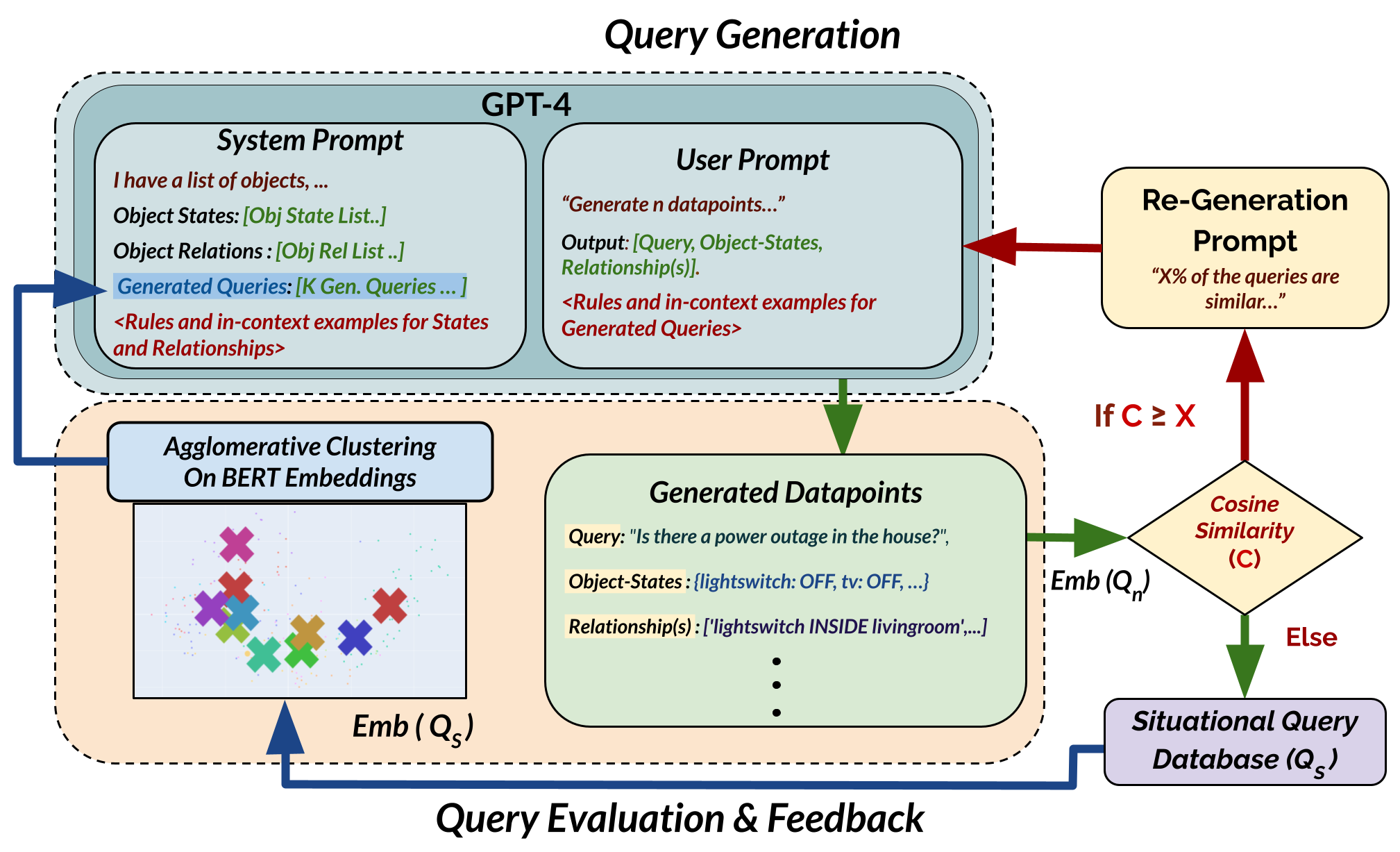} 
    \caption{\textbf{Prompt-Generate-Evaluate (PGE) Scheme}: We utilize an LLM (GPT-4) to generate \textit{situational datapoints} comprised of queries, corresponding consensus object states and relationships. The data generation occurs over \textbf{\textit{m}} iterations, refining LLM prompts after evaluating every batch of \textit{\textbf{n}} datapoints. For each batch, BERT embeddings of the $n$ queries are computed and denoted as $Emb(Q_{n})$. A cosine similarity $\mathds{C}$ is then computed between $Emb(Q_{n})$ and $Emb(\mathds{Q}_{S})$, the embeddings of existing queries in the \textit{Situational Query Database} $\mathds{Q}_S$. $\mathrm{C}$ is a threshold used to determine if the generated queries are sufficiently different from those in the database and if not, we continue \textit{conversation} with the LLM (indicated with the \textcolor{BrickRed}{red} arrows). Concurrently, we cluster $Emb(\mathds{Q}_{S})$ into $\textbf{k}$ categories, selecting the query nearest to the centroid in each cluster for feedback, labeled as ``Generated Queries'' in the System Prompt (denoted by \textcolor{NavyBlue}{blue} arrows). The ideal datapoint generation pathway is highlighted by the \textcolor{OliveGreen}{green} arrows.}
    \label{fig:gpt-4}
    \vspace{-0.6cm}
\end{figure*}

\section{Related Work}

\textbf{Embodied Question Answering (EQA)}:
EQA is a popular task \cite{eqasurvey, iqa, iqa2} that involves a robot attempting to answer a query by exploring and gathering information from its surroundings. It was introduced by Das et. al \cite{eqa} as a single-target question-answering dataset.
Yu et. al. \cite{mteqa} further extended this task to a multi-object setting, MT-EQA, where correspondences between multiple target objects present in the question itself need to be analyzed before arriving at an answer. Both these works use a template-based approach to generating questions, and this was deemed sufficient since the queries are objective in nature mentioning the objects of interest in their query itself. This however is a limitation, since a majority of queries tend to be subjective or situational in nature \cite{subjective_database}, especially in a household environment.

Sintean et. al in K-EQA \cite{knowledge-eqa} attempt to tackle this problem, by using scene-graph data and formal grammar to postulate queries. However, these queries are still limited to simple object grounding within a scene, and do not require any form of situational awareness about the \textit{readiness} of an environment. For instance, S-EQA considers queries like ``Is the living room ready for a movie night?'', that requires consensus on multiple objects and their states (TV: On, Popcorn: Cooked, On Table, etc.) to measure this readiness. A K-EQA query for the same example could be constructed as ``Is there a device for watching movies in the living room?'', which requires only scene knowledge about the availability of a TV, but does not require \textit{situational} knowledge about the various object-states involved in a movie night.

In our work, we utilize an LLM to generate situational queries and approximate human consensus on object states. Our queries are an abstraction above the scene-knowledge based K-EQA queries, requiring \textit{consensus knowledge} about multiple object \textit{states} for validation.

\textbf{LLMs for Query Generation}:
A recent trend has seen the emergence of Large Language Models (LLMs) such as GPT-3.5 \cite{chatgpt}, GPT-4 \cite{gpt4}, LLaMA \cite{LLaMA}, Palm \cite{palm} and Palm-e \cite{palm-e}. These models are well known for their commonsense reasoning capabilities \cite{llmsurv}, especially in the context of question-answering. In embodied AI, these capabilities have recently been leveraged to produce superior planning and decision-making strategies during navigation \cite{lgx, llmnav3, llmnav1, llmnav2}. 
In our case, we wish to generate situational queries for embodied question answering and ask if the common sense reasoning capabilities of LLMs can both generate such queries as well as the the consensus object-states needed for \textit{sufficiently} answering them.

\section{Situational EQA: Generation \& Validation}

Having motivated the problem of \textit{Situational EQA}, we now present a detailed description of our approach. We present three phases \emph{---} generating situational data using an LLM (GPT-4), annotating the data via a large-scale user survey to produce a S-EQA dataset, and finally evaluating it on the VirtualHome simulator. 

\subsection{\textbf{Situational Query Definition}}
\label{sec:Definition}
Situational queries in our work are defined as questions requiring the assessment of a collection of consensus objects and states to reach an existential `Yes/No' answer. For example, a query such as \textit{``Is the livingroom ready for movie night?''} has a Yes/No answer based on whether consensus objects like the TV, lights, etc. are in their appropriate state. Additionally, we want to ensure that the queries are related to a situation and are not simple in nature, i.e., the target objects and states are not referenced in the query itself. For example, ``\textit{Is the \underline{sofa} \underline{blue}?}'' \cite{eqa} or ``\textit{How many \underline{objects} can be used to hold rice?}'' \cite{knowledge-eqa} are simple queries, as they do not deal with a \textit{situation}, but rather pose a query referencing objects in the environment. In contrast, ``\textit{Is the dining area set up for dinner?}'' is situational, as it doesn't explicitly mention any objects, but requires the identification of \textit{relevant} objects (eg. \textit{table, napkins}), and a \textit{consensus} on what their states should be (\textit{open, present}) for the answer. Considering the LLM as a function $\mathcal{L}$ that generates queries $q$, and $\mathds{Q}$ to be a set of all possible queries generated by $\mathcal{L}$, we want \textit{situational} queries $\mathds{Q}_{S} \subset \mathds{Q}$, which satisfies the following conditions - 

\begin{itemize}
    \item \textbf{Abstraction}: The query must not directly reference any object, i.e., it must omit words from a set $\mathbb{W}$ containing words such as ``object'', ``table'', ``chair'', etc.
    \item \textbf{Contextual}: The query \textbf{requires} an assessment of relevant objects and their states collectively satisfy a scenario (e.g., “Is the dining area set up for dinner?”), requiring identification of appropriate objects and consensus on their expected states (e.g., table present, napkins, lighting, etc.).
    \item \textbf{Binary}: The answer to the query must be binary i.e., Yes/No, and not subjective in nature.
\end{itemize}

Formally, we can define $\mathds{Q}_{S}$ as:
\begin{align}
\mathds{Q}_{S} = \{& q \in \mathds{Q} \mid \text{Abstraction}(q, \mathds{W}) \nonumber \\
&\land \text{Contextual}(q) \land \text{Binary}(q) \}
\label{eq:definition}
\end{align}

This formulation ensures that $\mathds{Q}_{s}$ consists only of queries that are abstract, questioning in form, and expect a binary Yes/No response. We present these conditions as in-context rules to the LLM to curate it's responses.



\subsection{\textbf{Generating Situational Queries: S-EQA}}

We prompt the LLM to generate datapoints of the format - \textit{[Situational Query, Object-States, Object-Relationships]}. For example,  

\begin{quote}

\textit{\textbf{Situational Query}: \\ ``Is the bathroom ready for a shower?" \\
\textbf{Object-States} : [lightswitch: [`On'], towels: [`Present'], Soap: [`Present'] ...] \\
\textbf{Object-Relationships} : [lightswitch inside bathroom, towels inside bathroom, ... ]}

\end{quote}

The object states and relationships (such as `towels inside bathroom being present') being generated represent scene-graph requirements that the LLM believes need to be checked for the query to satisfy equation \ref{eq:definition}.  



\noindent\textbf{VirtualHome Setup:}
\label{sec: vhomesetup}
We use the VirtualHome \cite{virtualhome} simulator for defining ground truth objects, states and properties in our experiments. It contains 7 different household environments containing over 390 objects. Unlike other embodied simulators, VirtualHome is unique in that it contains several household objects including interactable ones, with easily \textit{modifiable} locations and states. The objects are also visually different upon modification, and this allows us to perform VQA later on to evaluate our dataset. 

All objects in VirtualHome have a set of state properties, which are either physical \textit{Open/Closed} (eg. for a Door), and/or \textit{On/Off} (such as lights) or \textit{Present/None} (such as Apple).
These objects also have a relationship with either the room or another object. For example, \textit{Apple INSIDE Fridge} and \textit{Fridge INSIDE livingroom} are both valid relationships. The simulator has 4 different rooms into which all objects are placed - [livingroom, bedroom, kitchen, bathroom]. We use these room markings to group and analyze the generated queries later on.


\begin{figure*}[t]
    \centering
    \begin{subfigure}{0.32\textwidth}
        \includegraphics[height=3.5cm,width=\linewidth]{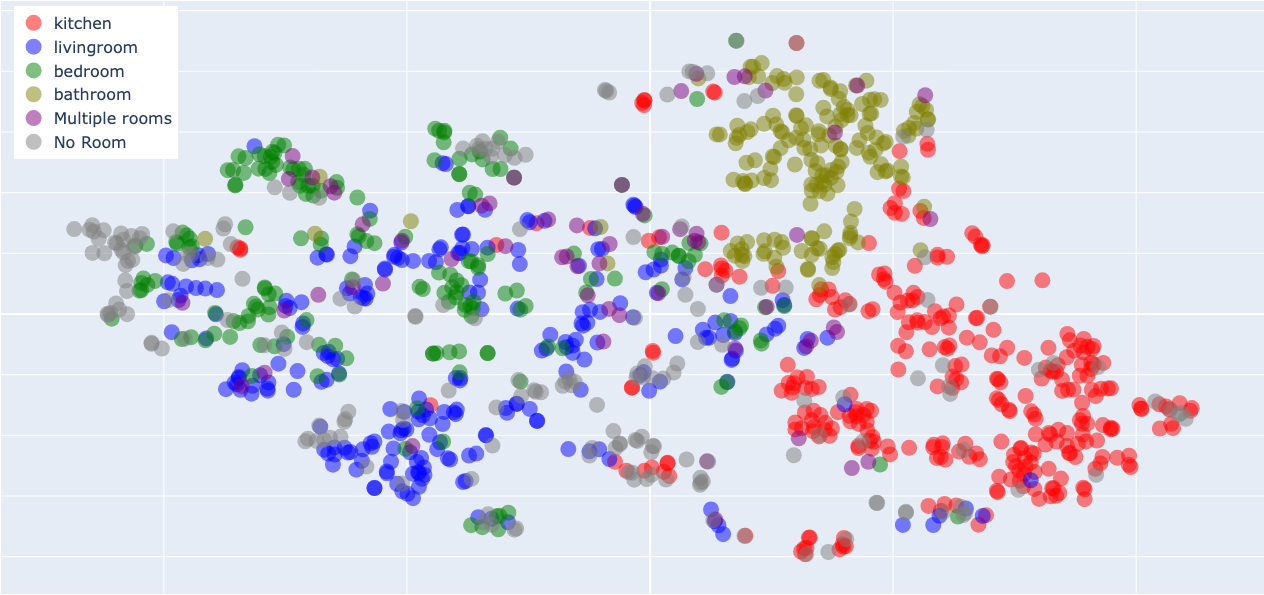}
        \caption{\textbf{Room Categories}}
        \label{fig:room_categories}
    \end{subfigure}
    \hfill
    \begin{subfigure}{0.33\textwidth}
        \includegraphics[height=3.5cm, width=\linewidth]{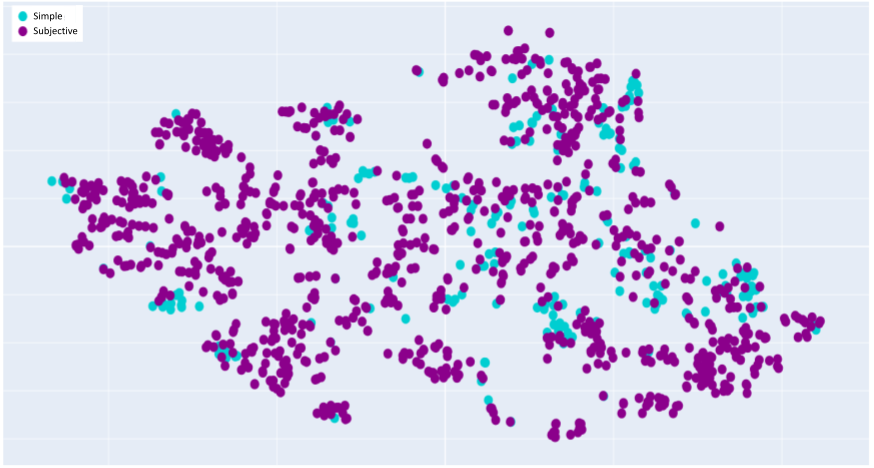}
        \caption{\textbf{Situational Categories}}
        \label{fig:comp_categories}
    \end{subfigure}
    \hfill
    \begin{subfigure}{0.32\textwidth}
        \includegraphics[height=3.5cm,width=\linewidth]{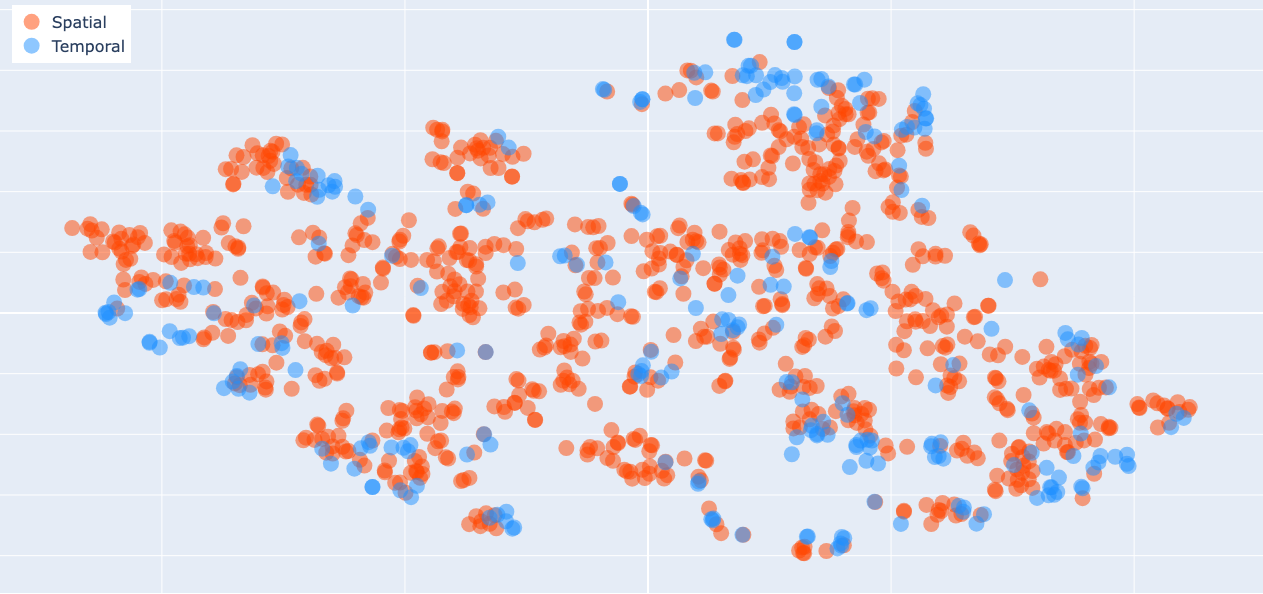}
        \caption{\textbf{Spatio-Temporal Categories}}
        \label{fig:temp_categories}
    \end{subfigure}
    \caption{\textbf{t-SNE plots of S-EQA}: We categorize and analyze the generated data in 3 aspects. For \textbf{Room categories}, we observe that most of the generated datapoints ($83.5\%$) belong to a particular room, suggesting that the LLM tends to create room-specific queries as opposed to multi or no-room scenarios. For \textbf{Situational Categories}, we observe that $82.20\%$ of the data generated is situational according to our definition. For \textbf{Spatio-Temporal Categories}, we look at whether the questions can be answered by only using the spatial positioning of objects and infer that $77.96\%$ of the queries are spatial.}
    \label{fig:tsne-plots}
    \vspace{-0.6cm}
\end{figure*}

\noindent\textbf{Prompt-Generate-Evaluate (PGE) Scheme:}
Figure \ref{fig:gpt-4} presents an overview of our \textit{Prompt-Generate-Evaluate} scheme for generating data. We analyze the LLM's generated datapoints to provide feedback for ensuring the data fits into our definition of situational queries, while also maintaining diversity.

The \textit{System Prompt} for our LLM contains a list of all object states and relationships, which in this case is the scene-graph provided by VirtualHome. We also provide additional rules and in-context examples to improve the generation of prompts. For instance, explicitly mentioning ``Do not conjure any new objects or states" ensures that the generated data does not contain objects not present or states that are not achievable in VirtualHome.
Additionally, we also provide a set of $k$ \textit{``Generated Queries"} as feedback to the system prompt. These queries are representative of what the LLM has already generated, and we encourage the LLM to generate queries different from these. 

Our \textit{User Prompt} asks to generate $n$ datapoints in the format specified earlier, and while providing a few manually annotated sample datapoints. Additionally, we provide some in-context rules to ensure that the generated queries are not too simplistic, not too ambiguous, and are also unique. For instance, we explicitly state `\textit{The situational query must not reference any object or even contain the word `object'.}' and `\textit{The query must have a Yes/No answer.}'. We also ask it to generate queries very different from what is already present in the $k$ \textit{Generated Queries} in the System Prompt and come up with `unique and creative household situations` with the object data provided.

\begin{figure}[h!]
    \centering
    \includegraphics[width=\linewidth]{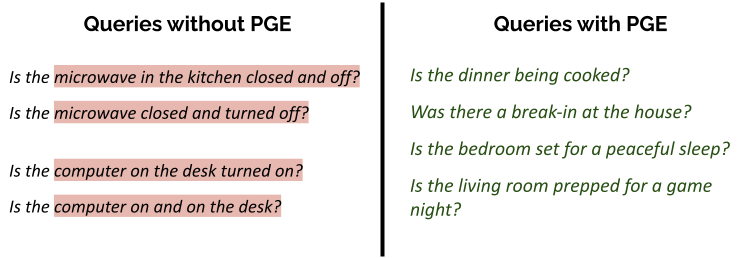}
    \caption{\textbf{Influence of PGE:} Notice the queries without PGE directly reference objects (microwave and computer) and are not situational. The queries are also just rephrased. PGE incorporates feedback measures to generate unique and diverse queries. Some queries may not hold general consensus on object states (such as \textit{Was there a break-in at the house?}), and these are filtered via human validation.}
    \label{fig:comparison}
    \vspace{-0.3cm}
\end{figure}

To ensure that our \textit{Situational Query Database} $\mathds{Q}_{S}$ contains diverse data points adhering to specifications presented in equation \ref{eq:definition}, we take the following two \textbf{feedback measures}:

\begin{itemize}
    \item \textbf{Generated Query Update:} After generating $n$ data points, we embed the situational queries using RoBERTa \cite{roberta} to get sentence embeddings $Emb(\mathds{Q}_{S})$. We perform agglomerative clustering \cite{agloclust} on $Emb(\mathds{Q}_{S})$ to accumulate them into $k$ groups of queries based on sentence similarity. We then pick the query closest to the centroid of each of these clusters to get $k$ queries that best represent the diversity in the generated data. Finally, we update the system prompt with these queries.
    
    \item \textbf{Continued Conversation:} Simultaneously, we compute the \textit{cosine similarity} $\mathds{C}$ of the $n$ generated query embeddings $Emb(Q_{n})$ with the queries already present in the \textit{Situational Query Database} $\mathds{Q}_{S}$. If $\mathds{C}$ is greater than a particular threshold ($X$), we continue the \textit{conversation} with the LLM, by presenting it with a re-generation prompt. We mention the percentage of similarity, and ask the LLM to re-generate a new batch of datapoints.
\end{itemize}

These \textit{feedback measures} ensure $\mathds{Q}_{S}$ contains unique situational queries that are not simple rewordings, while also encouraging diversity in the generated situations. Contrasting examples with and without PGE are illustrated in Fig. \ref{fig:comparison}.

The total number of datapoints generated $\mathds{Q}_{S}$ is always less than or equal to the product of the looping parameters $n$ and $m$, i.e., $\mathds{Q}_{S} \le n * m$. 
While our approach allows for the generation of potentially infinite datapoints, we observe a saturation point where the LLM starts to generate similar-looking queries (\textit{``Is the bathroom clean and dry?"} vs \textit{``Has the bathroom been cleaned today?"}). This is indeed a drawback of such a generative approach, and can be attributed to the abstract nature of the queries being generated, which do not pertain a particular object, but related to a household situation. Therefore, to effectively assess the utility of the data, we cut off generation at $2000$ data points. We validate this data via a large-scale MTurk user study, and present this as \textbf{S-EQA}, a dataset containing diverse and challenging situational queries along with consensus object data to help answer them. 

\noindent\textbf{Prompts:} We use the following prompts for PGE.

\noindent\textit{System Prompt:}
\begin{quote}
``I have a list of objects, and their states, and relationships in a household. \\
Object states are listed in the \texttt{OBJ\_STATE\_DICT} dictionary below. Each item has the format \texttt{OBJECT: [STATES]}. \\
STATES, IF PRESENT, can switch between \texttt{ON/OFF} and \texttt{OPEN/CLOSED}. Do not conjure any nrew objects or states. \\
\texttt{OBJ\_REL\_DICT} contains the initial relationships between the objects. These can be changed. For instance, \texttt{"apple INSIDE fridge"} is a valid relation. \\
\texttt{OBJ\_STATE\_DICT :-}
\{\dots\} \\
\texttt{OBJ\_REL\_DICT :-}
\{\dots\} \\
\texttt{GENERATED\_QUERIES :-}
\{\dots\} "
\end{quote}

\noindent\textit{User Prompt:}
\begin{quote}
`` Using \texttt{OBJ\_STATE\_DICT} and \texttt{OBJ\_REL\_DICT}, can you generate \{X\} potential questions, states and relationships that the user might ask about the environment? \\
These must be about a potential scenario, requiring situational awareness and a consensus on multiple object, their states and relationships. \\
The output should be lines of the form \texttt{[Question, Object-State Pairs, Relationships}].\\
Make sure you generate questions very different from those in \texttt{GENERATED\_QUERIES}. Get creative with the potential scenarios! \\
Make sure to use an \textbf{exhaustive set} of relationships and object-state pairs for each query.\\
The query must have a Yes/No answer. \\
The query must not directly reference any object or even contain the word `object'. "
\end{quote}

\noindent\textit{Re-Generation Prompt:}
\begin{quote}
``\{X\}\% of the questions are similar to what you've already generated earlier! Try again, give me \{Y\} more. \\
\texttt{QUERIES:}{\dots}"
\end{quote}

\subsection{\textbf{Validating S-EQA: t-SNE plots and M-Turk Annotation}}

Our objective with validating S-EQA is not just to annotate the queries, but to also measure the \textit{authenticity} of the generated data. We first visualize the queries in $\mathds{Q}_{S}$ using t-SNE plots to analyze their diversity. We then conduct a user study via Mechanical Turk \cite{mturk} to annotate the generated data.

\noindent\textbf{t-SNE Plots}:
We embed the generated queries with BERT and visualize them using t-SNE. They are then are categorized by Room (pick from [kitchen, living room, bedroom, bathroom, multi-room, no room]), Situational/Simple (Yes/No) and Spatio-Temporal (time-sensitivity). We observe that GPT-4 tends to generate queries specific to objects in a single room over multi-room or no-room ones like ``\textit{Is the house ready for sleeptime?}''. Further, we note that most queries ($82.2\%$) are classified as situational and not simple. Around $78\%$ are categorized as spatial queries (like ``\textit{Is the kitchen ready for cooking?}'') that refer to the current environment state, as opposed queries like ``\textit{Was someone in the kitchen today?}'' which require deeper situational understanding involving time. Overall, we observe that most queries adhere to the definition provided in Equation \ref{eq:definition}. 

\noindent\textbf{Room Categorization}:
Using the room properties obtained from VirtualHome (section \ref{sec: vhomesetup}), we group the questions into 6 categories - [kitchen, livingroom, bedroom, bathroom, multi-room, no room]. We infer from this plot that most questions being generated belong to one particular room, as opposed to multiple ones. This suggests that the LLM tends to generate situational queries that are usually room-specific, referring to multiple objects in a single area. As such, multi-room scenarios (like ``Is the house ready for sleeptime?") are less common in the generated data.

\noindent\textbf{Situational Categorization}:
Here, we use the LLM as a classifier to predict if the generated questions are situational according to our definition in \ref{sec:Definition}. We prompt and present in-context examples accordingly. We notice that $82.2\%$ of generated questions are classified as situational according to the LLM. As the LLM classifications have several false positives, this result motivates us to use human-validation to authenticate the dataset. 

\noindent\textbf{Spatio-Temporal Categorization}:
Here, we use the LLM to categorize the generated queries as spatial or temporal. Temporal questions rely on time-sensitive knowledge about a dynamic environment (Refer Fig. \ref{fig:temp_categories}). For instance, a query like ``\textit{Was someone sleeping in the bedroom today?}'' is temporal, as it requires knowledge of past environmental states to be answerable. Conversely, a query like ``\textit{Is the kitchen ready for cooking?}'' is spatial, as it requires the agent to gauge only current states. We observe that $77.96\%$ of the generated questions are Spatial ones, showing that PGE prefers modeling spatial relations over temporal ones.

\noindent\textbf{MTurk Validation:}
\label{sec:human-eval}
We conduct a large-scale Mechanical Turk \cite{mturk} study to annotate the generated data and evaluate the authenticity of real-world queries. The annotators are chosen from the US-East region, with at least a $95\%$ approval rate on previous work.

Annotators are shown a generated query, along with consensus object-states and relationships. Given options of \textit{Yes}, \textit{No}, and \textit{Cannot Answer}, a worker's capacity to give a definitive response, i.e., choosing \textit{Yes} or \textit{No}, serves as an indicator of whether the generated data is in alignment with general user consensus. For instance, the query ``Is the house ready for sleeptime?", if agreed upon by workers when shown images of dimly-lit rooms, closed doors, and windows, validates the data point.

We obtain a low `Cannot Answer' percentage of $2.74\%$ among the $5\times2K = 10K$ participants. As such, a high $97.26\%$ percentage of generated queries being answerable strongly indicates that the LLM is capable of generating object-states and queries that are a good representation of human consensus.

Subsequently, we conduct a second \textbf{comparison study} with the same workers, where they are presented with an image of a room accompanied by a situational S-EQA query (For eg. \textit{Is someone working in the bedroom?} with top-views of the bedroom) and, separately, images of objects paired with consensus queries formulated based on their state (For eg., \textit{Is the \underline{Computer} \underline{On}?} with an image of a computer being on).
We sample additional `Yes/No' questions the SimVQA \cite{simvqa} dataset to even out the number of situational and consensus queries. An improved annotator performance in answering consensus queries over situational ones would indicate S-EQA's \textit{utility} in breaking down subjective queries into simpler consensus queries for improved question-answering. We discuss this later in the results section.

In both our MTurk experiments, we measure average reviewer agreement across 5 users on each question to obtain ground truth labels on the data. For instance, if a question has $3$ \textit{Yes's}, and $2$ \textit{No's}, we label the ground truth as `Yes'. On the other hand, if even one of the annotations is `Cannot Answer', we label the human-annotated output by the same.

\begin{figure}
    \centering
    \includegraphics[width=\linewidth]{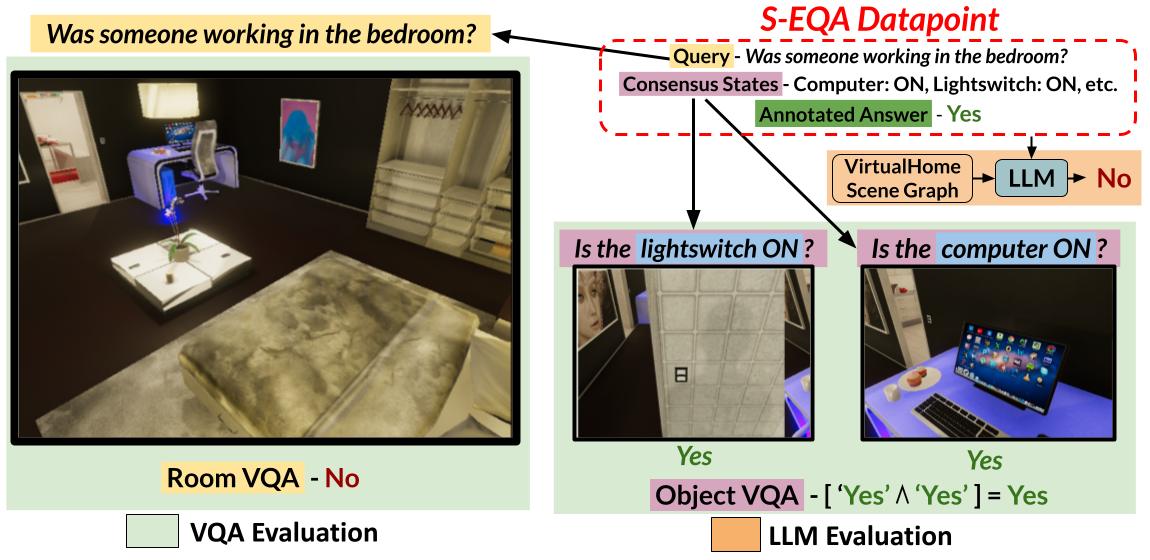}
    \caption{\textbf{S-EQA Simulator Eval}: We evaluate S-EQA generated from VirtualHome using the LLM with Scene Graphs (\textcolor{YellowOrange}{Orange}) and VQA (\textcolor{YellowGreen}{Green}). In this example, the situational query asks if someone was working in the bedroom, to which the annotation was positive (Yes). 
    For the LLM Eval, we pass the entire VirtualHome scene graph with modified consensus states along with the query as input to the LLM. Note the negative response of the LLM here, despite it generating valid consensus states. This mismatch is present in $\textbf{46.8\%}$ of cases, reflecting poor LLM answering capability. For VQA Eval, we query various VLMs with Room and Object images (See Table \ref{tab:vqa-analysis}). Note the simpler Object VQA queries that are answered correctly, suggesting breaking down situational queries into consensus ones can simplify the task via indirect answering. 
    }
    \label{fig:vqa}
    \vspace{-0.5cm}
\end{figure}

\section{Situational EQA: Evaluation}
Figure \ref{fig:vqa} presents an overview of our evaluation of the simulator-based dataset, and Figure \ref{fig:rwsetup} presents our real-world setup for evaluation. We evaluate queries both via VQA as well as by using an LLM. Further, LLM answers are qualitatively studied via reasoning to infer situational awareness.

\subsection{\textbf{VQA Evaluation:}} 

We perform VQA on images taken from VirtualHome with situational and consensus queries. The VirtualHome simulator is unique in enabling this, as it allows us to modify consensus object states pertaining to subjective queries such that they also visually change upon modifying states. 

\noindent\textbf{Situational $\rightarrow$ Consensus Queries}:
Consensus object states associated with the generated situational queries are used to formulate binary objective questions for evaluation. These are of the form - 
\begin{quote}
\centering
    \textit{Is the $<$OBJECT$>$ $<$STATE$>$?}
\end{quote}

For instance, for a query $q:$`\textit{Was someone working in the bedroom?}', the consensus queries would be `\textit{Is the Computer On?}' and `\textit{Is the Lightswitch On?}'.

We present VQA models with both situational and consensus queries, and compare the results.

\noindent\textbf{Room VQA (Situational Query)}: Here, we present the  VQA model with multiple top-view images of a relevant room as input, and ask it to answer the associated situational query with Yes/No/Cannot Answer. The assumption here is that the agent might have access to overhead security cameras which it could use to answer the queries. 
Most of our queries are associated with one particular room, which is often part of the query itself. During our experimentation, we observe a bias in model performance when the name of the room itself is present in the query. For instance, if the query contained `living room', and the image was that of the living room, the answer would always be yes, irrespective of the actual context of the query. To mitigate this, we replace any mention of a room in the query with the generic term - ``\textit{this place}".

\noindent\textbf{Object VQA (Consensus Query)}: 
We provide the VQA model with these consensus queries along with images of the corresponding objects in their designated states.

For both \textbf{Room VQA} and \textbf{Object VQA}, we measure the accuracy and F1-scores of prediction across all the images, using the human-verified answers as ground truth.
We use a threshold of $0.5$ on the accuracy and F1 scores to determine if the query has successfully been answered. 
Additionally, we also compute the \textbf{Joint Accuracy} ($\mathrm{J}_{A}$), and \textbf{Joint F1} ($\mathrm{J}_{F1}$) scores which is a success case when either of the Room or Object VQA are successful. Formally, let \( R \) be the accuracy or F1 result for Room VQA and \( O \) be the same result for Object VQA, where \( R, O \in \{0, 1\} \). \( R = 1 \) indicates a match with the ground truth for Room VQA, and similarly, \( O = 1 \) indicates a match for Object VQA.

$\mathrm{J}$ is then defined as:
\[
\mathrm{J}(R, O) = 
\begin{cases} 
1 & \text{if } R = 1 \text{ or } O = 1 \\
0 & \text{otherwise}
\end{cases}
\]

$\mathrm{J}_{a}$ and $\mathrm{J}_{F1}$ give us an upper bound on the performance of VQA models. For $\mathrm{J}_{F1}$, we set a threshold of $0.5$ above which the score would be deemed satisfactory.

\subsection{\textbf{LLM Evaluation:}} 
Motivated by K-EQA \cite{knowledge-eqa}, we utilize scene graphs provided by VirtualHome to develop an LLM-based action planner.

For S-EQA, we have situational queries generated along with consensus object states and relationships. VirtualHome provides us with a \textit{fully observable scene graph} for this, i.e., the states and relationships of all objects are known. 

Our aim is to measure how well the LLM is able to approximate the human consensus behind answering situational queries. For this, we first ask the LLM to provide a Yes/No answer to each of the the datapoints generated. We then measure the correlation between the LLM and human annotated answer.
Further, we also ask the LLM to reason about its answers for a subjective analysis.


\subsection{\textbf{Real World Experiments:}}  

\begin{figure*}
    \centering
    \includegraphics[width=\linewidth]{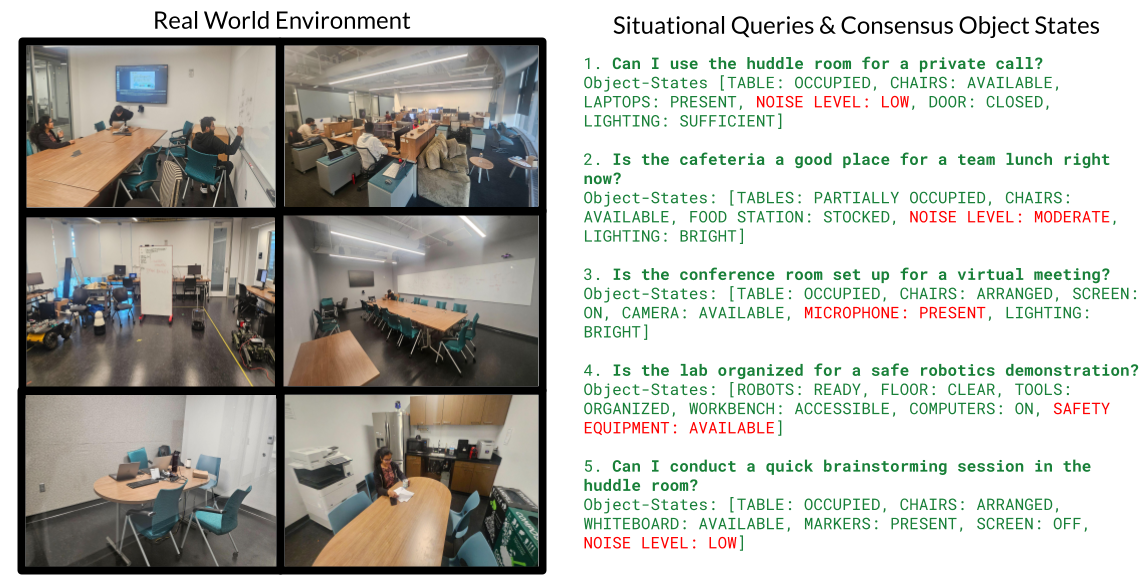}
    \caption{\textbf{Real World Setup}: Since we do not have a scene graph readily available in the real world, we pass the images as context to GPT-4o. On the left are images showing various situations in different areas in our environment. On the right are examples of the generated situational queries and consensus object state data. Note the object-states marked in red, indicating \textbf{LLM hallucination} generating consensus information that is outside the scope of the data provided. VirtualHome datapoints did not have this issue since the scene graph was explicitly given. Directly using a VLM to generate this data seems to produce such hallucinations.}
    \label{fig:rwsetup}
    \vspace{-0.5cm}
\end{figure*}

Unlike VirtualHome, where a fully observable scene graph is readily available, the real-world environment does not have object states in a predefined format. Instead, we gather images from various locations in our environment which is used as context for GPT-4V to generate situational datapoints. 
A representative subset of scenes and datapoints being generated are illustrated in Fig \ref{fig:rwsetup}. We aimed to create diverse situations including group meetings, robot operations, cafeteria related events across various locations in our lab environment.  

The VLM predicts Yes/No responses for each datapoint. These are then compared against human-annotated ground truth responses to assess alignment. We then analyze failure cases where the LLM misinterprets missing information or hallucinates object states, showcasing the difficulty in accurately comprehending situational queries in the real world.

\section{Results and Inferences}
GPT-4 \cite{gpt4} is our LLM of choice to both generate and evaluate situational queries in our experiments. Our ablations with GPT-3.5 showed poor quality in terms of generation, with many outputs not fitting the specified data format. 

\subsection{Simulator Setup}
We first program the object states and relationships from the generated data points into the VirtualHome environment. This also includes manually filtering out a small subset of bad states which are not feasible. For instance, \textit{`computer INSIDE fridge'} is not a valid relationship for us to set, and the simulator prevents this.

For VQA evaluation, we require images of the room and corresponding consensus objects set to their appropriate states. We gather room images by considering the 4 corners of each room, with a camera pointing downwards towards the center. Choosing images in this way allows us to achieve maximum coverage of objects present.

Then for object images, we use the first-person view of a human placed in the environment. We consider the human to be our robot agent here. They are instructed to walk to each of the objects present in a S-EQA datapoint, where we grab an image when they are present in front of them. This approach for capturing images proved to be simpler to implement over simulator camera placement and gives us a visual perspective of an agent grabbing an object.

\subsection{VQA Results}
We use BLIP-2 \cite{blip-2}, BLIP \cite{blip} and ViLT \cite{vilt} to evaluate S-EQA, presented in Table \ref{tab:vqa-analysis} below.

\begin{table}[ht]
    \centering
    \begin{tabular}{llcc}
    \hline
    & \textbf{VQA Model} & \textbf{Accuracy \%} $\uparrow$ & \textbf{F1-Score \%} $\uparrow$ \\
    \hline
    \multirow{3}{*}{\rotatebox[origin=c]{90}{\textbf{Room}}} & BLIP-2 \cite{blip-2} &$\textbf{42.31}$ & $\textbf{52.51}$ \\
    & BLIP \cite{blip} & $41.82$ & $49.63$ \\
    & ViLT \cite{vilt} & $39.80$ & $42.31$ \\
    \hline
    \multirow{3}{*}{\rotatebox[origin=c]{90}{\textbf{Object}}} & BLIP-2 \cite{blip-2} & $\textbf{58.26}$ & $\textbf{68.75}$ \\
    & BLIP \cite{blip} & $51.33$ & $63.75$ \\
    & ViLT \cite{vilt} & $47.27$ & $58.31$ \\
    \hline
    \multirow{3}{*}{\rotatebox[origin=c]{90}{\textbf{Joint}}} & BLIP-2 \cite{blip-2} & $\textbf{89.6}$ & $82.5$ \\
    & BLIP \cite{blip} & $83.29$ & $\textbf{83.75}$ \\
    & ViLT \cite{vilt} & $72.91$ & $80.35$ \\
    \hline
    \end{tabular}
    \caption{VQA performance on Room and Object VQA.}
    \label{tab:vqa-analysis}
    \vspace{-0.3cm}
\end{table}

Observe the higher Object-VQA accuracies ($15.31\%$ on average) and F1 scores over Room-VQA across each approach. This suggests that answering situational queries by looking at images at a room level is significantly harder than answering queries about the states of target objects. Our \textbf{comparison study} with MTurk (Refer \ref{sec:human-eval}) further corroborates this result, with $74.3\%$ of users finding consensus queries to be simpler to answer over situational ones. This suggests that our approach of breaking down situational queries into simpler consensus ones could be used to indirectly answer difficult situational queries.

The high values on joint scores $\mathrm{J}$ indicate an upper bound on performance.
While still better than RoomVQA, ObjectVQA models have mediocre performance overall, despite the queries being straightforward in nature. This can be attributed to inference on synthetic images from VirtualHome which these models have not been trained on, opening an avenue for future work.

Language - Question length, re-classification percentage, Plot curve of average length of dataset questions.  
VQA - Room VQA and Object VQA
\begin{figure}[h]
    \centering
    \includegraphics[width=\linewidth]{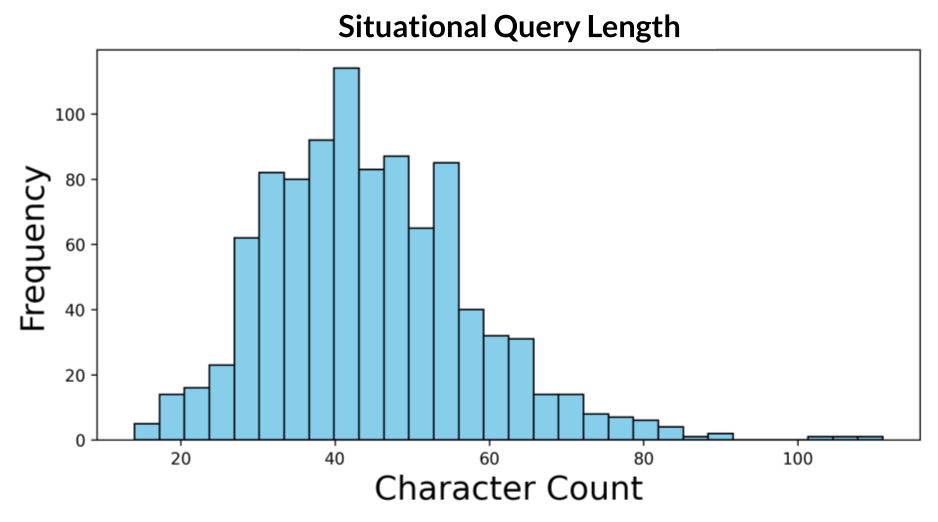}
    \caption{Distribution of Query Lengths}
    \label{fig:lenplot}
    \vspace{-0.3cm}
\end{figure}

Figure \ref{fig:lenplot} provides a distribution of the query lengths in S-EQA. Note that this distribution resembles a Gaussian, with the median character count of queries being around $40.2$, and word count being $7.7$. This indicates that most generated queries in S-EQA are of an ideal length, with GPT-4 producing queries with concise scenario-specific language as opposed to an accumulation of multiple object states like MT-EQA \cite{mteqa}.

\subsection{LLM Eval Results}

We obtain a low correlation of $46.2\%$ between GPT-4 generated answers and the human-annotated ones. This is despite the LLM having complete access to the entire scene graph which it used earlier to generate situational queries.

Furthermore, the low `Cannot Answer' percentage of $2.74\%$ on the annotated responses shows that most queries and consensus data generated did indeed make sense to most users. We can infer from these two values that while GPT-4 is good at generating situational queries and the potential consensus data needed to answer them, it is poor at answering it's own generated queries. To understand this further, we prompt for reasoning as follows:-

\noindent\textbf{LLM Reasoning:}
To analyze the poor LLM performance on query answering, we reason about the generated answers by prompting GPT-4 to provide reasoning as follows: 

\begin{quote}
    \textit{$<$Situational Query and Object Data$>$ \\
    Your answer is: $<$LLM Answer (Yes/No)$>$. \\
    Can you provide a brief reason for your answer focusing only on the object states and relationships provided?}
\end{quote}

Upon inspecting the generated reasoning, we notice that GPT-4 often goes against the consensus while predicting the answer. For example, consider the following datapoint -

\begin{quote}

\textit{\textbf{Query}: ``Is the living room prepared for a movie night?" \\
\textbf{Object-States} : [tv: [`ON'], lights: [`OFF'], remotecontrol: [`ON'], ...] \\
\textbf{Object-Relationships} : [tv INSIDE livingroom, lights INSIDE livingroom, remotecontrol INSIDE livingroom, ...] \\
\textbf{LLM Answer}: `NO' \\}

\end{quote}

The reasoning generated for the LLM Answer is - 
\begin{quote}
    \textit{The lights are off which is suitable for a movie night, but the TV is on which is not suitable for movie watching as it may create distractions.}
\end{quote}

GPT-4 suggests that the TV being ON may create distractions for movie watching and hence responds with No. However, a more plausible situation would be that the user would watch the movie on the TV itself, and that the living room is indeed prepared for a movie night.

On the flip side, since our object states are limited to the ones present in VirtualHome, we are inclined to believe that the quality of the output depends on the richness of the object set. In this example, for instance, additional object information such as \textit{DVD: [`Present']} and \textit{Popcorn ON Table} would improve the consensus of a movie night about to happen. Since such objects do not exist in VirtualHome, it is reasonable to assume that GPT-4's explanation is sufficient.


These results suggest that while GPT-4 is indeed capable of producing situational queries and \textit{authentic} consensus information, its displays \textbf{poor commonsense reasoning} while answering such queries.

\begin{figure*}
    \includegraphics[width=\linewidth]{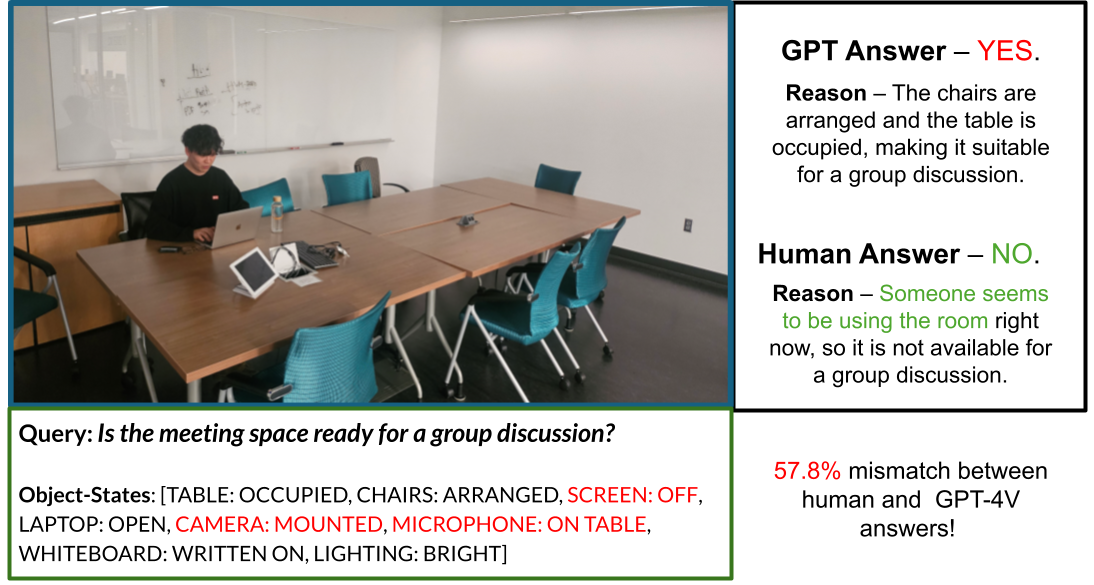}
    \caption{\textbf{LLM Reasoning}: On our real world data, we use PGE to generate situational query and object data. We then annotate these queries manually via an in-house user study, and then compare their answers with the LLM generated ones. Note the \textbf{unnatural reasoning} showcased by the LLM when asked to explain its generated answers. In this instance, there is already someone that seems to be using the conference room, and hence it is not ready for a group meeting. But the LLM ignores this fact and and suggests that the table being occupied makes it suitable for a meeting. This mismatch is on $57.8\%$ of the generated data, suggesting that the LLM exhibits \textbf{poor commonsense reasoning} on our task.}
    \label{fig:rw_inspect}
\end{figure*}

\subsection{Real World Experiments: Results and Discussion}

We experiment with situational queries in our real-world lab environment, assuming the presence of an assistive workplace robot agent. We gather a total of $10$ images across various situations, corresponding to $6$ different locations in our workplace, and feed these into PGE as visual context for GPT-4V. We then generate $100$ datapoints using PGE and manually annotate them.

During annotation, we note several hallucinations in the object states that did not occur earlier when the scene graph was explicitly defined in VirtualHome. Some instances of this are highlighted in red in Figure \ref{fig:rwsetup}. For example, note the noise level indication is present as an object state in many cases, but can not be discerned by looking at the image. Ghost objects such as \textit{microphones} and \textit{safety equipment} also appear in the generated object-states. We infer here that a \textit{grounded} scene graph, like what VirtualHome provides is important to avoid poor data generation.

Evaluating the generated datapoints with GPT-4 presents a mismatch of $57.8\%$, a similar trend as that with the VirtualHome data, and solidifies our claim that LLM generated binary answers are misaligned with human consensus. 

Similar to what was done with VirtualHome, we conduct LLM based reasoning on the queries generated. An example of this is illustrated in Fig. \ref{fig:rw_inspect}. Note the counterintuitive explanation that GPT gives when asked if the meeting space is ready for a group discussion. Despite the table being occupied with someone present in the room, who also appears to be using their laptop (Object State includes Laptop:OPEN), the reasoning is completely off. We observe many such cases in the data that we gather, where LLM answers and the justification provided goes against human consensus.

LLMs have been widely adopted in robotics for their strong commonsense reasoning, enabling significant performance gains across embodied tasks \cite{lgx, palm-e, mu2023embodiedgpt}. These models have contributed to the rise of foundation models for robotics, where a common paradigm involves LLMs predicting task outputs or structured action plans, which agents then execute with varying affordances \cite{saycan}. Given these successes, one would expect LLMs to exhibit similar strengths in our task.

However, we observe a counterintuitive outcome—LLMs perform poorly in commonsense reasoning. While this may stem from the inherent \textit{subjectivity} of situational queries, it also reveals GPT-4V’s limited situational awareness, even when provided with visual cues. Additionally, hallucinations in object states likely degrade downstream query performance, highlighting how even minor inaccuracies in state identification can significantly impact task execution. This contrasts with prior work where LLMs have demonstrated strong object reasoning, suggesting that situational awareness presents a distinct challenge. Future work will look at improving LLM-based situational reasoning by employing more robust multimodal alignment schemes via image captioning, memory-aware mechanisms, or explicit uncertainty modeling to mitigate state hallucinations and enhance task reliability.

\section{Conclusion, Limitations and Future Work}
We introduce and tackle the novel and challenging problem of Embodied Question Answering with Situational Queries \emph{---} \textbf{S-EQA}. Unlike prior work that dealt with simple objective queries that directly reference target objects, situational queries necessitate an approximation of a \textit{consensus} on multiple object states. Given the challenge in gathering such queries from the real world, we rely on commonsense knowledge from an LLM, GPT-4 to develop a Prompt-Generate-Evaluate (PGE) framework for posing unique situational queries and corresponding consensus object information.
In the VirtualHome simulator, we use PGE to generate 2K unique queries, and annotate them via a large scale MTurk user study with $10K$ participants. The high answerability ($97.26\%$) lets us infer that LLMs are good at generating situational queries and corresponding consensus data.


However, using GPT-4 to answer the simulator queries shows a low correlation with the human annotations ($46.2\%$); we infer that \textbf{LLMs are good at generating situational data but poor at answering them}. When asked to explain its answers, we observe the \textbf{LLM's poor commonsense capability}, exhibiting counterintuitive reasoning. This suggests that LLMs show poor situational awareness.

Finally, we discuss challenges with transferring our generation and evaluation schemes to a real world lab environment. Since a grounded scene graph isn't directly available for use in the real-world, we provide images of various locations as context for query generation, and observe several unintended hallucinations with \textit{ghost} object states as consensus. Future work will focus on limiting these by first gathering structured scene graphs using existing visual map building techniques. 




{\small
\bibliographystyle{IEEEtran}
\bibliography{refs}

\begin{thebibliography}{10}
\providecommand{\url}[1]{#1}
\csname url@samestyle\endcsname
\providecommand{\newblock}{\relax}
\providecommand{\bibinfo}[2]{#2}
\providecommand{\BIBentrySTDinterwordspacing}{\spaceskip=0pt\relax}
\providecommand{\BIBentryALTinterwordstretchfactor}{4}
\providecommand{\BIBentryALTinterwordspacing}{\spaceskip=\fontdimen2\font plus
\BIBentryALTinterwordstretchfactor\fontdimen3\font minus \fontdimen4\font\relax}
\providecommand{\BIBforeignlanguage}[2]{{%
\expandafter\ifx\csname l@#1\endcsname\relax
\typeout{** WARNING: IEEEtran.bst: No hyphenation pattern has been}%
\typeout{** loaded for the language `#1'. Using the pattern for}%
\typeout{** the default language instead.}%
\else
\language=\csname l@#1\endcsname
\fi
#2}}
\providecommand{\BIBdecl}{\relax}
\BIBdecl

\bibitem{virtualhome}
X.~Puig, K.~Ra, M.~Boben, J.~Li, T.~Wang, S.~Fidler, and A.~Torralba, ``Virtualhome: Simulating household activities via programs,'' in \emph{Proceedings of the IEEE Conference on Computer Vision and Pattern Recognition}, 2018, pp. 8494--8502.

\bibitem{eqa}
A.~Das, S.~Datta, G.~Gkioxari, S.~Lee, D.~Parikh, and D.~Batra, ``Embodied question answering,'' in \emph{Proceedings of the IEEE conference on computer vision and pattern recognition}, 2018, pp. 1--10.

\bibitem{eqa2}
H.~Luo, G.~Lin, F.~Shen, X.~Huang, Y.~Yao, and H.~Shen, ``Robust-eqa: robust learning for embodied question answering with noisy labels,'' \emph{IEEE Transactions on Neural Networks and Learning Systems}, 2023.

\bibitem{eqa3}
D.~Gao, R.~Wang, Z.~Bai, and X.~Chen, ``Env-qa: A video question answering benchmark for comprehensive understanding of dynamic environments,'' in \emph{Proceedings of the IEEE/CVF International Conference on Computer Vision}, 2021, pp. 1675--1685.

\bibitem{mteqa}
L.~Yu, X.~Chen, G.~Gkioxari, M.~Bansal, T.~L. Berg, and D.~Batra, ``Multi-target embodied question answering,'' in \emph{Proceedings of the IEEE/CVF Conference on Computer Vision and Pattern Recognition}, 2019, pp. 6309--6318.

\bibitem{iqa}
D.~Gordon, A.~Kembhavi, M.~Rastegari, J.~Redmon, D.~Fox, and A.~Farhadi, ``Iqa: Visual question answering in interactive environments,'' in \emph{Proceedings of the IEEE conference on computer vision and pattern recognition}, 2018, pp. 4089--4098.

\bibitem{knowledge-eqa}
S.~Tan, M.~Ge, D.~Guo, H.~Liu, and F.~Sun, ``Knowledge-based embodied question answering,'' \emph{IEEE Transactions on Pattern Analysis and Machine Intelligence}, 2023.

\bibitem{mturk}
``{A}mazon {M}echanical {T}urk --- mturk.com,'' \url{https://www.mturk.com/}.

\bibitem{eqasurvey}
J.~Duan, S.~Yu, H.~L. Tan, H.~Zhu, and C.~Tan, ``A survey of embodied ai: From simulators to research tasks,'' \emph{IEEE Transactions on Emerging Topics in Computational Intelligence}, vol.~6, no.~2, pp. 230--244, 2022.

\bibitem{iqa2}
Q.~Sima, S.~Tan, H.~Liu, F.~Sun, W.~Xu, and L.~Fu, ``Embodied referring expression for manipulation question answering in interactive environment,'' in \emph{2023 IEEE International Conference on Robotics and Automation (ICRA)}.\hskip 1em plus 0.5em minus 0.4em\relax IEEE, 2023, pp. 7635--7641.

\bibitem{subjective_database}
Y.~Li, A.~X. Feng, J.~Li, S.~Mumick, A.~Halevy, V.~Li, and W.-C. Tan, ``Subjective databases,'' \emph{arXiv preprint arXiv:1902.09661}, 2019.

\bibitem{chatgpt}
OpenAI, ``Language models are unsupervised multitask learners,'' \emph{OpenAI Blog}, vol.~23, no.~6, 6 2020, \url{https://openai.com/blog/chatgpt}.

\bibitem{gpt4}
------, ``Gpt-4 technical report,'' 2023.

\bibitem{LLaMA}
H.~Touvron, T.~Lavril, G.~Izacard, X.~Martinet, M.-A. Lachaux, T.~Lacroix, B.~Rozi{\`e}re, N.~Goyal, E.~Hambro, F.~Azhar \emph{et~al.}, ``Llama: Open and efficient foundation language models,'' \emph{arXiv preprint arXiv:2302.13971}, 2023.

\bibitem{palm}
A.~Chowdhery, S.~Narang, J.~Devlin, M.~Bosma, G.~Mishra, A.~Roberts, P.~Barham, H.~W. Chung, C.~Sutton, S.~Gehrmann \emph{et~al.}, ``Palm: Scaling language modeling with pathways,'' \emph{arXiv preprint arXiv:2204.02311}, 2022.

\bibitem{palm-e}
D.~Driess, F.~Xia, M.~S. Sajjadi, C.~Lynch, A.~Chowdhery, B.~Ichter, A.~Wahid, J.~Tompson, Q.~Vuong, T.~Yu \emph{et~al.}, ``Palm-e: An embodied multimodal language model,'' \emph{arXiv preprint arXiv:2303.03378}, 2023.

\bibitem{llmsurv}
J.~Huang and K.~C.-C. Chang, ``Towards reasoning in large language models: A survey,'' \emph{arXiv preprint arXiv:2212.10403}, 2022.

\bibitem{lgx}
V.~S. Dorbala, J.~F. Mullen~Jr, and D.~Manocha, ``Can an embodied agent find your" cat-shaped mug"? llm-based zero-shot object navigation,'' \emph{arXiv preprint arXiv:2303.03480}, 2023.

\bibitem{llmnav3}
G.~Zhou, Y.~Hong, and Q.~Wu, ``Navgpt: Explicit reasoning in vision-and-language navigation with large language models,'' \emph{arXiv preprint arXiv:2305.16986}, 2023.

\bibitem{llmnav1}
D.~Shah, B.~Osi{\'n}ski, S.~Levine \emph{et~al.}, ``Lm-nav: Robotic navigation with large pre-trained models of language, vision, and action,'' in \emph{Conference on Robot Learning}.\hskip 1em plus 0.5em minus 0.4em\relax PMLR, 2023, pp. 492--504.

\bibitem{llmnav2}
C.~H. Song, J.~Wu, C.~Washington, B.~M. Sadler, W.-L. Chao, and Y.~Su, ``Llm-planner: Few-shot grounded planning for embodied agents with large language models,'' \emph{arXiv preprint arXiv:2212.04088}, 2022.

\bibitem{roberta}
Y.~Liu, M.~Ott, N.~Goyal, J.~Du, M.~Joshi, D.~Chen, O.~Levy, M.~Lewis, L.~Zettlemoyer, and V.~Stoyanov, ``Roberta: A robustly optimized bert pretraining approach,'' \emph{arXiv preprint arXiv:1907.11692}, 2019.

\bibitem{agloclust}
D.~M{\"u}llner, ``Modern hierarchical, agglomerative clustering algorithms,'' \emph{arXiv preprint arXiv:1109.2378}, 2011.

\bibitem{simvqa}
P.~Cascante-Bonilla, H.~Wu, L.~Wang, R.~S. Feris, and V.~Ordonez, ``Simvqa: Exploring simulated environments for visual question answering,'' in \emph{Proceedings of the IEEE/CVF Conference on Computer Vision and Pattern Recognition}, 2022, pp. 5056--5066.

\bibitem{blip-2}
J.~Li, D.~Li, S.~Savarese, and S.~Hoi, ``Blip-2: Bootstrapping language-image pre-training with frozen image encoders and large language models,'' \emph{arXiv preprint arXiv:2301.12597}, 2023.

\bibitem{blip}
J.~Li, D.~Li, C.~Xiong, and S.~Hoi, ``Blip: Bootstrapping language-image pre-training for unified vision-language understanding and generation,'' in \emph{International Conference on Machine Learning}.\hskip 1em plus 0.5em minus 0.4em\relax PMLR, 2022, pp. 12\,888--12\,900.

\bibitem{vilt}
W.~Kim, B.~Son, and I.~Kim, ``Vilt: Vision-and-language transformer without convolution or region supervision,'' in \emph{International Conference on Machine Learning}.\hskip 1em plus 0.5em minus 0.4em\relax PMLR, 2021, pp. 5583--5594.

\bibitem{mu2023embodiedgpt}
Y.~Mu, Q.~Zhang, M.~Hu, W.~Wang, M.~Ding, J.~Jin, B.~Wang, J.~Dai, Y.~Qiao, and P.~Luo, ``Embodiedgpt: Vision-language pre-training via embodied chain of thought,'' \emph{Advances in Neural Information Processing Systems}, vol.~36, pp. 25\,081--25\,094, 2023.

\bibitem{saycan}
M.~Ahn, A.~Brohan, N.~Brown, Y.~Chebotar, O.~Cortes, B.~David, C.~Finn, C.~Fu, K.~Gopalakrishnan, K.~Hausman \emph{et~al.}, ``Do as i can, not as i say: Grounding language in robotic affordances,'' \emph{arXiv preprint arXiv:2204.01691}, 2022.

\end{thebibliography}

}

\end{document}